\documentclass[letterpaper, 10 pt, conference]{ieeeconf}
\IEEEoverridecommandlockouts
\usepackage{cite}
\usepackage{amsmath,amssymb,amsfonts}
\usepackage{algorithmic}
\usepackage{graphicx}
\usepackage{siunitx}
\usepackage[utf8]{inputenc}
\usepackage{textcomp}
\usepackage[dvipsnames]{xcolor}
\usepackage[T1]{fontenc}
\usepackage{amsfonts}
\usepackage{xcolor}

\usepackage{float}
\let\labelindent\relax
\usepackage{enumitem}
\usepackage{epsfig}
\usepackage{epstopdf}
\usepackage[fleqn]{nccmath}
\usepackage[linesnumbered, ruled, vlined]{algorithm2e}
\usepackage[colorlinks,urlcolor=blue]{hyperref}

\def\BibTeX{{\rm B\kern-.05em{\sc i\kern-.025em b}\kern-.08em
    T\kern-.1667em\lower.7ex\hbox{E}\kern-.125emX}}
\newcommand\figref{Fig.~\ref}

\makeatletter
\newcommand*{\rom}[1]{\expandafter\@slowromancap\romannumeral #1@}
\makeatother

\SetCommentSty{mycommfont}

\begin{document}

\title{\LARGE \bf Risk-Averse MPC via Visual-Inertial Input and Recurrent Networks \\ for Online Collision Avoidance}

\author{Alexander Schperberg$^{*1}$, Kenny Chen$^{*2}$, Stephanie Tsuei$^{3}$, Michael Jewett$^{1}$, Joshua Hooks$^{1}$, \\ Stefano Soatto$^{3}$, Ankur Mehta$^{2}$, and Dennis Hong$^{1}$
\thanks{$^{1}$A. Schperberg, M. Jewett, J. Hooks, and D. Hong are with the Robotics and Mechanisms Laboratory, Department of Mechanical and Aerospace Engineering. {\tt\small \{aschperberg28, mbjewett318, hooksj, dennishong\}@ucla.edu}}
\thanks{$^{2}$K. Chen and A. Mehta are with the Laboratory for Embedded Machines and Ubiquitous Robots, Department of Electrical and Computer Engineering. {\tt\small \{kennyjchen, mehtank\}@ucla.edu}}
\thanks{$^{3}$Stephanie Tsuei and S. Soatto are with the UCLA Vision Lab, Department of Computer Science. {\tt\small \{stephanietsuei, soatto\}@ucla.edu}}
\thanks{All labs are affiliated with the University of California, Los Angeles in Los Angeles, CA 90095, USA}
\thanks{\textit{Asterisk (*) denotes equal contribution}}
\thanks{\newline \textbf{© 2020 IEEE. Personal use of this material is permitted. Permission from IEEE must be
obtained for all other uses, in any current or future media, including
reprinting/republishing this material for advertising or promotional purposes, creating new
collective works, for resale or redistribution to servers or lists, or reuse of any copyrighted
component of this work in other works.}}
}

\maketitle

%============================================================
% Abstract and Keywords
%============================================================

\begin{abstract}
In this paper, we propose an online path planning architecture that extends the model predictive control (MPC) formulation to consider future location uncertainties for safer navigation through cluttered environments. Our algorithm combines an object detection pipeline with a recurrent neural network (RNN) which infers the covariance of state estimates through each step of our MPC's finite time horizon. The RNN model is trained on a dataset that comprises of robot and landmark poses generated from camera images and inertial measurement unit (IMU) readings via a state-of-the-art visual-inertial odometry framework. To detect and extract object locations for avoidance, we use a custom-trained convolutional neural network model in conjunction with a feature extractor to retrieve 3D centroid and radii boundaries of nearby obstacles. The robustness of our methods is validated on complex quadruped robot dynamics and can be generally applied to most robotic platforms, demonstrating autonomous behaviors that can plan fast and collision-free paths towards a goal point.
\end{abstract}

%============================================================
% Introduction
%============================================================

\section{Introduction}

\begin{figure*}[!t]
    \centering
    \includegraphics[width=\textwidth]{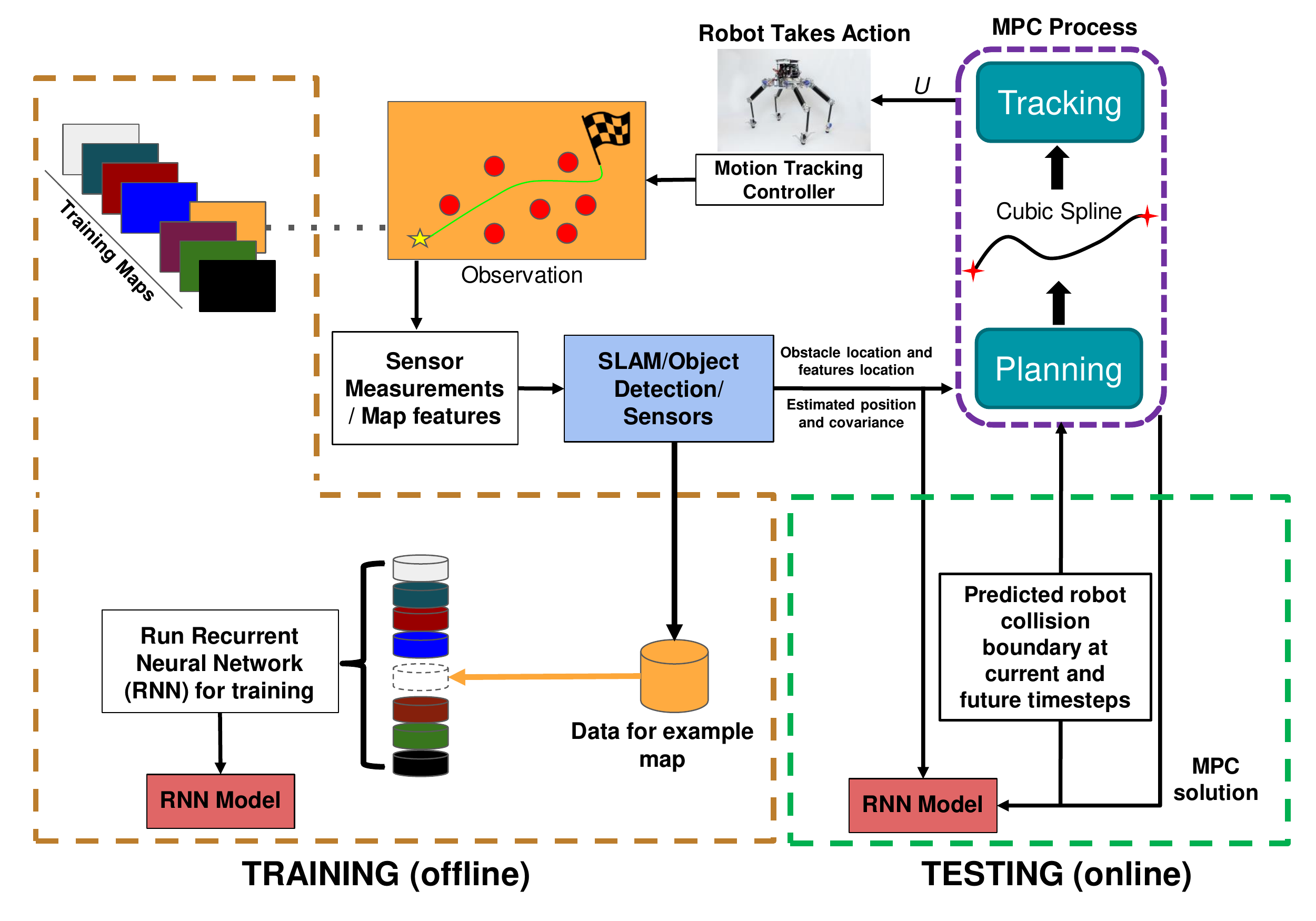}
    \caption{\textbf{Architecture Overview.} This figure demonstrates the training and testing procedures of our method. In training, we first select different maps, where obstacles in each map are randomly distributed. A simulation where the robot moves from an initial to a goal position is executed on this map. At each timestep an observation is taken (e.g., camera or on-board sensor data). These measurements are used as the input to our SLAM/Object Detection/Sensors system, which estimate the current position and uncertainty in position of the robot, and also location and size of obstacles. MPC accounts for this information and produces outputs entered into our motion tracking controller. For every map at every timestep, the current observations, state position, and positional uncertainty (among other variables outlined in Section~\ref{methods:rnn}) are entered into a large database to produce our RNN model. Lastly, in the testing phase, RNNs can predict the positional uncertainty (which provide our collision boundaries) of the robot at future timesteps of the MPC prediction horizon.}
    \label{fig_scram}
\end{figure*}

For robots to truly become a viable option for unmanned tasks such as search and rescue operations and unknown environmental exploration, autonomous and safe trajectory planning must be considered as a fundamental design goal during algorithmic design and systems integration. Necessary requirements for robots to autonomously perform such complex tasks include, but are not limited to, online low-level feedback controls, localization, vision, motion planning, high-level reasoning, and reasoning under uncertainty. Currently, all individual components are well-developed, but integrating multiple pieces together into a single system, especially for environments that are not well-known, has proven to be a daunting challenge because of issues related to robustness \cite{robustness}. For example, simultaneous planning, localization, and mapping (SPLAM, or "Active SLAM") is an active area of research that attempts to satisfy some of these requirements. The main challenges to Active SLAM consist of planning under uncertainty in an acceptable amount of time, bridging the gap between sensor data ("semantic mapping"), and ensuring that it is robust enough for a complex platform. Because of these challenges, there are many works addressing a subset of Active SLAM, namely simultaneous localization and planning \cite{slap} or works addressing SLAM, such as \cite{orb-slam, jones_visual-inertial_2011, lsd-slam}. Other Active SLAM works also use very simple research platforms \cite{ActiveSLAM2} or were only tested in simulation \cite{ActiveSLAM}.

There are two common frameworks to address the problem of planning under uncertainty, which are explicitly modeling the posterior distributions in a Bayesian setting \cite{prob_planning} or using a partially-observable Markov decision process (POMDP) \cite{pomdp}. However, the Bayesian setting is only computationally tractable for the simplest cases (e.g. Gaussian prior and Gaussian observations) and POMDPs suffer from the curse of history and dimensionality and do not sufficiently model an agent's future intent \cite{rnn_uncertainty}. Recent work addressing Active SLAM using POMDPs also either lacks mapping capability \cite{slap} or requires an inordinate amount of computational resources \cite{pomdpThesis}. As a consequence, there is a need for new path planning architectures for unknown and uncertain environments that addresses the concerns of belief space planning or provides alternative methods that can be ubiquitously applied on most robotic systems.

To resolve the above issues, we propose a multifaceted approach that uses model predictive control (MPC), SLAM\footnote{In this paper, "SLAM" includes visual-inertial odometry with sparse mapping in addition to algorithms that produce denser maps.}, and recurrent neural network (RNN) algorithms to address the problem of Active SLAM and account for uncertainties in both current and future robot positions. Our architecture is based on MPC because MPC operates online, continually satisfies the dynamic state of the robot over a prediction horizon $N$, and naturally offsets estimation errors \cite{relativefootstepMPC}. The MPC is augmented to be "risk-averse" by considering uncertainty in position from timestep $k$ to $k+N$. This uncertainty is inferred by an RNN, which has been demonstrated to handle time series data, account for temporal factors that directly affect predictions, have shown promise in modeling complex interactions between agents and their environment \cite{rnn_uncertainty,rnn_1} and previously applied to MPC but for industrial processes \cite{lanzetti2019recurrent}. Our RNN model is trained on the positional covariance estimations of a visual-inertial odometry (VIO) system taking readings from an inertial measurement unit (IMU) and camera data as input. By considering the current and future positional uncertainties in the MPC optimization problem, our method can solve for more optimal control actions at each timestep. To facilitate object avoidance, we additionally incorporate an object detection pipeline that uses a deep convolutional neural network (CNN) to recognize obstacles and a feature detector with RGB and depth images to estimate the distance and size of nearby obstacles. We show that by using a trained RNN model to infer positional uncertainties at future timesteps, a robot can demonstrate more evasive behavior to better guarantee collision avoidance without becoming too conservative. Our linearized path planning framework is applied and tested on a complex quadruped robot, which demonstrates our algorithm's robustness and efficiency in computation, showing the feasibility of extending our work to a wide range of robotic platforms.

\subsection*{Summary of Our Contributions}

(1) We evaluate the feasibility of an online end-to-end path planner that unifies MPC, SLAM, RNN, and an object detector using CNNs to generate paths for unknown and uncertain environments using a non-linear programming solver.

%that account for the following assumptions: (a) the robot's location in relation to the surrounding map and obstacles is uncertain; (b) the environment is unexplored \textit{a priori} and on-board or external sensor measurements have uncertainty; (c) calculating optimized states must be performed online and in small time intervals while accounting for a dynamic environment; and (d) the dynamics of the robot must be satisfied.

(2) We verify that our quadrupedal robot, ALPHRED \cite{hooks2020ALPHRED}, avoids collisions and computes a shorter trajectory while maintaining safety using our method as compared to a more conservative and naive planner.

%A legged robot was chosen over more simple robotic systems to showcase that our linearized framework is still robust. 

(3) We propose a novel use of RNNs to estimate positional uncertainties at all future timesteps of the MPC's prediction horizon.

(4) We integrate all components into a high-fidelity simulation using the quadruped dynamics of ALPHRED (Figure \ref{gazebo_sim}). Additionally, we test all components individually either online or offline using hardware (Figure \ref{fig_alphred}). 

In the following sections, we will explicitly refer to the simulation or hardware data. The main difference between the model of ALPHRED in simulation versus hardware is that in simulation the model is equipped with an idealistic RGB + dense depth Microsoft Kinect camera, while the actual hardware is equipped with Intel's Realsense D435i. The idealistic camera publishes both RGB and dense depth images at arbitrarily fast speeds while the RealSense publishes RGB images at 30Hz and dense depth images at only 2Hz.

%============================================================
% Methods
%============================================================

\section{Methods}

In this section, we provide an overview of our architecture and how our risk-averse MPC propagates uncertainty through its finite time horizon trajectory. In Section~\ref{methods:architecture}, we provide a high-level overview of our path planning algorithm. In Section~\ref{methods:mpc}, we describe our MPC's mathematical framework for planning and tracking. In Section~\ref{methods:mpc-constraints}, we describe the constraints used in our cost functions. In Section~\ref{methods:obstacle}, we describe our object detection system using CNNs and keypoint detection on RGB and depth images, and finally in Section~\ref{methods:rnn}, we detail our RNN training and inference procedures (utilizing our SLAM algorithm) for predicting future positional uncertainties used to create collision boundaries.

\subsection{Architecture Overview}
\label{methods:architecture}

Our path planner is formulated as an MPC optimization problem using a non-linear programming solver \cite{Casadi}. We divide our MPC framework into a planning phase and a tracking phase, with different cost functions for each. In the planning phase, the goal of our MPC is to create waypoints that move a robot closer to a desired position while detecting and avoiding obstacles through measurement updates. Specifically, the object detection algorithm feeds the MPC with the position and size of surrounding obstacles, while the positional uncertainty of the robot at all future timesteps in the MPC prediction horizon is inferred by RNNs. In the tracking phase, we discretize the generated path into segments of fixed temporal length using a cubic polynomial to create a smooth reference trajectory. MPC is used to track this reference trajectory and outputs our desired planar velocity values ($v_{d}$ and $\dot{\psi}_d$). These velocity values are used by our motion tracking controller to generate stable footstep trajectories. Note that dividing our MPC formulation into two phases facilitates lower computation time, and allows for separate control on waypoint generation and creation of custom reference trajectories if desired (see Algorithm \ref{ra_mpc}, \figref{fig_scram}, or our accompanied video\footnote{\url{https://youtu.be/td4K55Tj-U8}} for a general overview of our path planning architecture).

\begin{algorithm}[!ht]
	\SetAlgoLined
	\DontPrintSemicolon
	Initialize state $X$, control $U$, $dt_{plan}$, $dt_{track}$, horizon $N$, robot collision boundary $r_{\Sigma_{k:k+N}}$, timestep $k$; $bboxes$ = bounding boxes \\
	\BlankLine
	\tcp{Planning Phase (waypoints to goal)}
	\While{$ \| X_{curr} - X_{goal} \|_2 > 0$}{
	
        $X, U_{sols} \leftarrow \text{MPC}(X_{curr}, X_{goal}, r_{\Sigma_{k:k+N}})$ \\
        
        $X_{ref}, U_{ref} \leftarrow \text{CubicSpline}(X, U_{sols})$ \\
        
        $[pixel_x, pixel_y]_{1:f} \leftarrow \text{FeatureExtractor}(\text{RGB})$ \\
        
        $bboxes \leftarrow \text{CNN}(\text{RGB})$ \\
        
        $[x, y, z]_{1:l} \leftarrow \text{ObjectDetector}(\text{RGB-D}, [pixel_x, pixel_y]_{1:f}, bboxes)$ \\
        
        $X_{curr} \leftarrow \text{RobotEstimator}(U, \text{IMU}, \text{joint encoders})$ \\
        
        $r_{\Sigma_{k:k+N}} \leftarrow \text{RNN}(X_{sols}, [x, y, z]_{1:l})$ \\
        
        \BlankLine
        \tcp{Tracking Phase (follow $X_{ref}$ and $U_{ref}$)}
        \While{$dt \leq dt_{plan}$}{
            
            $U \leftarrow \text{MPC}(X_{curr}, X_{ref(dt)}, U_{ref(dt)})$ \\
            
            $X_{curr} \leftarrow \text{MotionTrackingController}(U)$ \\
            
            $dt \mathrel{{+}{=}} dt_{track}$ \\
        }
        $dt = 0$ \\
	}
	\caption{Risk-Averse MPC}
	\label{ra_mpc}
\end{algorithm}

\subsection{General MPC Formulation}
\label{methods:mpc}
\subsubsection{Planning Phase}
MPC in the planning phase has the following time-invariant linear discretized model:
\begin{equation}\label{motionmodel} 
    {f\left(X_{k}, U_{k}\right)=X_{k+1}=A X_{k}+BU_{k}+w_{k}} 
\end{equation}

\noindent where $X=\left[\begin{array}{l}{x}, \ {y} \end{array}\right]^\top$ represents our state variables (planar waypoint position), and $U=\left[\begin{array}{l}{v_{x}}, {v_{y}}\end{array}\right]^\top$ represents our control variables (planar velocity). We also initialized our state and control variables to zero before run-time.

Because we have a motion tracking controller to incorporate robot dynamics (see Section \ref{experimental results:A}), our A and B matrices can assume a simple point mass:

\begin{center}
    $A$ = $\left[\begin{array}{cc}
    {1} & {0} \\
    {0} & {1} \\
    \end{array}\right]$, $B$ = $\left[\begin{array}{cc}
    {dt_{plan}} & {0} \\
    {0} & {dt_{plan}} \\
    \end{array}\right]$ \\
\end{center}

\noindent where $dt_{plan}$ is the time between taking proprioceptive and exteroceptive sensor measurements (e.g., RGB-D images and odometer readings), and $w_k$ represents a non-unit variance random Gaussian noise ($w_k \sim \mathcal{N}(0, \sigma^2$), where $\sigma$ represents the standard deviation of planar velocity).

The goal of our cost function in the planning phase (equation \eqref{objfunction1}) is to find the optimal control value that minimizes the distance from the current and predicted states ($X_{k=0\rightarrow N}$) to the goal state ($X_{goal}$) -- where $X_{k=0}$ is given by the results of localization. Note, that we use $\hat{U}_{k}$ instead of $U_{k}$ in our cost function to represent the inclusion of a slack decision variable, $\epsilon$ (the slack variable has no role in our discretized model equation, but does affect the cost function through $R$ - see \ref{methods:mpc-constraints}), so that $\hat{U}=\left[\begin{array}{l}{v_{x}}, {v_{y}}, {\epsilon} \end{array}\right]^\top$.

\begin{equation}
    \begin{split}
        \min_{U_{k:k+N}} \sum_{k=0}^{N} \left(X_{k+1}-X^{goal}\right)^{\top}Q\left(X_{k+1}-X^{goal}\right) + \hat{U}_{k}^{\top}R\hat{U}_{k}\\
    \end{split}
\label{objfunction1}
\end{equation}
\begin{center}
s.t.  \rom{1}, \rom{2}, \rom{3}, \rom{5} (see Table~\ref{MPC_Constraints})
\end{center}

\subsubsection{Tracking Phase}
MPC in the tracking phase has the following time-invariant linear discretized model:
\begin{equation}\label{motionmodel2} 
    {f\left(X_{k}, U_{k}\right)=X_{k+1}=A X_{k}+BU_{k}}
\end{equation} 
\noindent where $X=\left[\begin{array}{l}{x}, \ {y}, \ {\psi} \ \end{array}\right]^\top$ represents our state variables (desired planar position and yaw or heading angle), and $U=\left[\begin{array}{l}{v_{x}}, {v_{y}}, {\dot\psi}\end{array}\right]^\top$ represents our control variables (desired planar velocity and yaw rate). Matrices $A$ and $B$ are the same as shown in \eqref{motionmodel}, except for an additional row/column for yaw and yaw rate.

The goal of our cost function in the tracking phase (equation \eqref{objfunction2}) is to output desired planar velocity and yaw rate ($v_{d}$ and $\dot{\psi}_d$) values that follow a reference trajectory. 

\begin{equation}
\label{objfunction2}
    \begin{split}
        \min_{U_{k:k+N}} \sum_{k=0}^{N} \left(X_{k}-X_{k}^{ref}\right)^{\top}Q\left(X_{k}-X_{k}^{ref}\right)\\
        + \left(U_{k}-U_{k}^{ref}\right)^{\top}R\left(U_{k}-U_{k}^{ref}\right) \\
    \end{split}
\end{equation}

\begin{center}
s.t. \rom{1}, \rom{2}, \rom{3}, \rom{4} (see Table~\ref{MPC_Constraints})   
\end{center}

\noindent $X^{ref}$ and $U^{ref}$ are obtained by cubic interpolation (equation \eqref{cubic_polynomial}) with end points specified by the MPC planning phase from $X_{k}...X_{k+2}$ and $U_{k}...U_{k+2}$ (the reason we discretize to ${k+2}$ instead of $k+1$ is to ensure there are enough reference points for MPC to "look-ahead").

%The cubic polynomial is discretized into segments %of length $dt_{track}$.
\begin{equation}\label{cubic_polynomial}
    \begin{split}
    X^{ref}(t), U^{ref}(t)=a_{0}+a_{1} t+a_{2} t^{2}+a_{3} t^{3}, \\ \quad a_{i}=f\left(dt_{track}, X_{k}, {U}_{k}, X_{k+2}, {U}_{k+2}\right)
    \end{split}
\end{equation}

\subsection{MPC Constraints}
\label{methods:mpc-constraints}

\subsubsection{Constraints \rom{1} - \rom{4}} Constraint \rom{1} represents multiple shooting constraints which facilitate solving non-linear programs \cite{MS_constraints}. The limits on state variables (i.e., map constraints), and control variables (limits on velocity) are represented by Constraint \rom{2} (note that the slack decision variable should be set as $0 \leq \epsilon$). If there is apparent jerk during path planning, it may be necessary to include Constraint \rom{3}, where $\alpha^{limit}$ represents the limit on acceleration ($a_{x}$, $a_{y}$, and $\ddot{\psi}$) and $U$ represents velocity ($v_{x}, v_{y}$ and $\dot{\psi}$). Orienting the robot along the planned trajectory can be achieved using Constraint \rom{4}, and setting the limit on $v_{y}$ to be much smaller than the limit on $v_{x}$ (which points directly along the path) in the body frame. Because MPC outputs velocities in the inertial reference frame ($irf$), a rotation matrix is required to transform these velocities into the correct frame of reference.

\subsubsection{Constraint V - Collision Boundary with Slack Variable}

Our obstacle avoidance constraints are given by Constraint \rom{5}, which ensure that the collision boundary of the robot does not collide with detected obstacles (note, that because these constraints are updated at every timestep $dt_{plan}$, moving obstacles can also be considered). $x_{o_i}$ and $y_{o_i}$ represent the $x$ and $y$ center positions of all obstacles detected by the robot ($i\rightarrow M$: where $M$ is the number of obstacles currently in range). $x_k$ and $y_k$ represent the $x$ and $y$ positions of the robot from timestep $k$ to timestep $k+N$ (future positions can be received from the MPC solution). $r_{\Sigma_{k}}$ represents the radius of the collision boundary of the robot, and $r_{o_i}$ represents the radius of the collision boundary of the obstacle. The collision boundary radius of the robot is calculated by using the major axis of the covariance uncertainty ellipse ($\Sigma$) estimated from RNN and is added to the radius or size of the robot itself (thus, we assume a more conservative collision boundary, which, combined with the slack variable---see below---provides some tolerance to ensure the planner does not fail while at the same time lowering the probability of collision). Note, from timestep $k$ to timestep $k+N$, $\Sigma_{k\rightarrow N}$ is predicted by our RNN (see Section~\ref{methods:rnn}). 

Without a slack variable and because of sensor measurement noise, the measured state of the robot may suddenly find itself very near or slightly inside the collision boundary of the obstacle and cause the solver to fail. To accommodate for this issue, Constraint \rom{5} includes a slack variable to allow for some degree of constraint violation in our optimization problem. In other words, we effectively separate the covariance into a constraint and slack variable, where we tune the confidence attributed to the posterior estimate and break it into a nominal estimate, and a controllable slack parameter. Specifically, slack can be tuned through the $R$ weighing matrix, where a high cost for $\epsilon$ will ensure that the majority of solutions will not violate Constraint \rom{5}, while lower values allow for greater violation (the cost on the slack variable is largely dependent on user-experience during implementation).

\begin{table}[!t]
\centering
    \caption{Model Predictive Control Constraints}
    {\renewcommand{\arraystretch}{2}%
    \begin{tabular}{||r c||} 
    \hline
    No. & Constraint \\
    \hline\hline
    
    I & $X_{k+1}-f\left(X_{k}, U_{k}\right)=0$ \\
    
    II & $X^{limit} \geq |X_{k}|, \ U^{limit} \geq |U_{k}|$ \\
    
    III & $\mathrm{\alpha}^{limit} \geq|\left[\mathrm{U}_{\mathrm{k}+1}-\mathrm{U}_{\mathrm{k}}\right]| / \mathrm{d} \mathrm{t}$ \\
    
    IV & $\left[\begin{array}{c}
            {v_{x}^{limit}} \\
            {v_{y}^{limit}}
            \end{array}\right]_{body} \geq \left | \left[\begin{array}{cc}
            {\cos \theta_{k}} & {\sin \theta_{k}} \\
            {-\sin \theta_{k}} & {\cos \theta_{k}}
            \end{array}\right]\left[\begin{array}{l}
            {v_{x}} \\
            {v_{y}}
            \end{array}\right]_{irf} \right |$ \\
    
    V & $-\sqrt{\left(x_{k}-x_{o_i}\right)^{2}+\left(y_{k}-y_{o_i}\right)^{2}}+r_{\Sigma_{_k}}+r_{o_i} - \epsilon \leq 0$ \\
    
    \hline
    \end{tabular}} \quad
    \label{MPC_Constraints}
\end{table}

\subsection{Obstacle Detection}
\label{methods:obstacle}

\subsubsection{Convolutional Neural Networks}
\begin{figure}[!t]
    \centering
    \includegraphics[width=0.95\columnwidth]{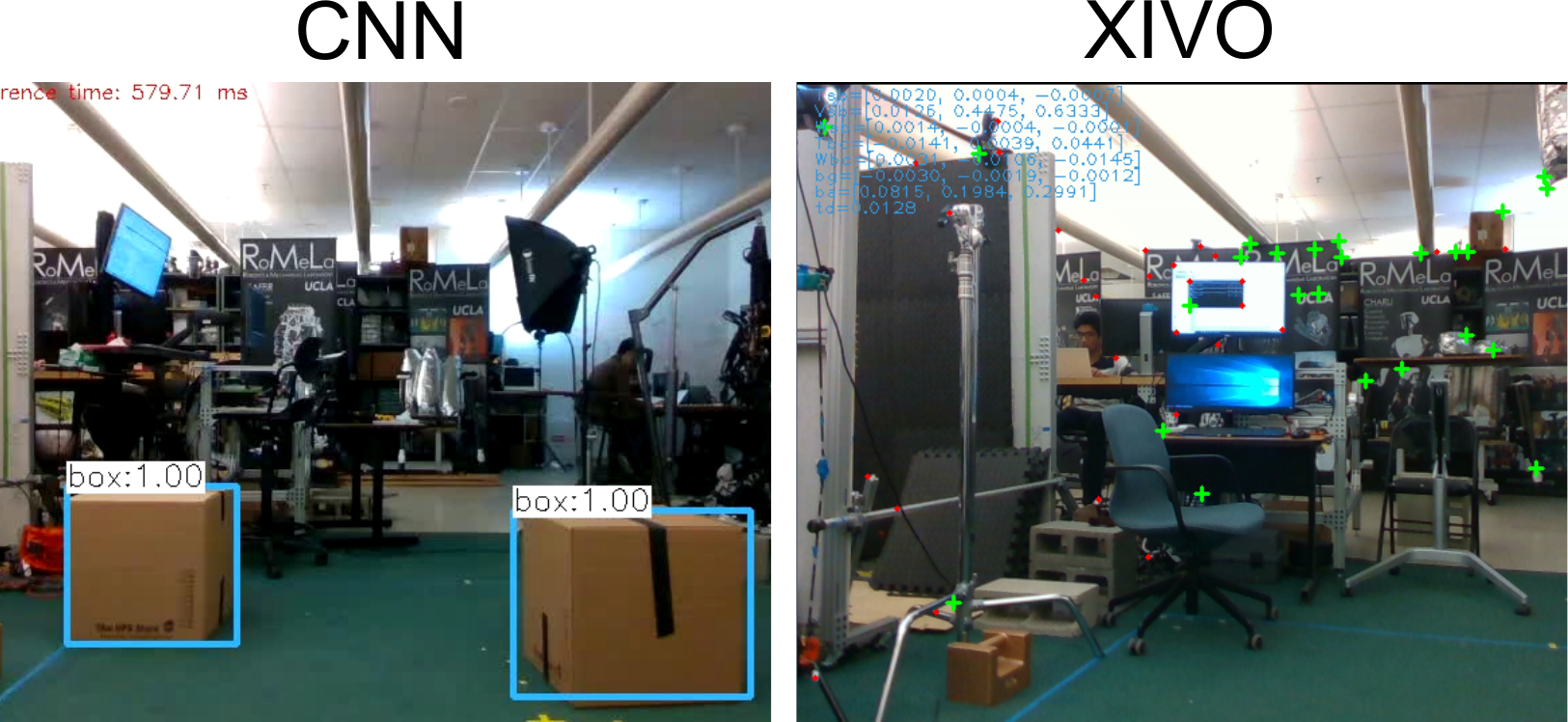}
    \caption{\textbf{Example Module Outputs.} \textit{Left}: An example output image of our trained object detector using a custom-trained convolutional neural network model. We used the YOLOv3 \cite{yolov3} architecture with default initialized weights for fast training and inference. \textit{Right}: Inlier (green +) and outlier tracks (red *) produced by XIVO on data collected from the Intel Realsense D435i.}
    \label{cnn_box}
\end{figure}

To support the avoidance of incoming obstacles as our robot traverses the environment, we use a custom-trained CNN model for real-time object detection. More specifically, using Redmon et al.'s YOLOv3 \cite{yolov3} fast CNN architecture because of its maturity (although other methods could be used, such as \cite{xiao-semantic-mapping}), we trained two custom models. One localized brown boxes within an RGB camera frame using 1,500 hand-labeled images and the other localized black $1m$ x $1m$ boxes in a mostly empty Gazebo environment using 300 images. Weights were initialized using YOLOv3's default weights and trained for 5,200 epochs using stochastic gradient descent with a batch size of 64, momentum of 0.9, and learning rate of 0.001 for both models. We validated our model on labeled data withheld from the training data and we verified empirically that our object detector could successfully draw tight bounding boxes around our brown boxes (Fig.~\ref{cnn_box}).

\subsubsection{From Bounding Boxes to 3D Obstacles}
\label{methods:obstacle-keypoint}

For our end-to-end Gazebo simulation, we implemented simple classical feature detection over the simulated RGB and dense depth images to transform the bounding boxes from the object detector into useful 3D obstacles for the motion planner. The scheme described below assumes that features are cubes and that ALPHRED is directly facing all existing boxes. It is executed only once, at the beginning of the simulation. Note that instead of fully addressing the semantic mapping problem, we use simple placeholder computer vision components designed for our specific test scenarios; for now, we only utilize SLAM as training data for the RNN.

First, ORB features \cite{orb-features} are extracted. Let $x_p$ and $y_p$ be the pixel coordinates of a single feature, and $Z_c$ be its depth. Then, let $g_{sb} = (R_{sb}, T_{sb})$ be the body-to-spatial transformation, $g_{bc} = (R_{bc}, T_{bc})$ be the camera-to-body transformation, and $K$ be the intrinsics matrix. Then, the position of the feature in the spatial frame, $X_s$ (a 3x1 vector) can be calculated as:

\begin{equation}
\begin{aligned}
\begin{bmatrix} x_c \\ y_c \\ 1 \end{bmatrix} &= K^{-1} \begin{bmatrix} x_p \\ y_p \\ 1 \end{bmatrix} \\
X_c &= Z_c \begin{bmatrix} x_c \\ y_c \\ 1 \end{bmatrix} \\
X_b &= R_{bc} X_c + T_{bc} \\
X_s &= R_{sb} X_b + T_{sb}
\end{aligned}
\end{equation}

Next, for each bounding box captured by the CNN object detection process, we determine which features are in each bounding box. The size of the box is the maximum distance (in meters) between any two points. Half of that size then becomes the "radius" of the obstacle's collision boundary (the MPC assumes that the collision boundary are circles).

We note that our classical feature detection approach is computationally efficient but also simple (i.e., not as robust). For example, most features exist near corners, where rounding errors could lead to a very different depth value. For the purpose of our end-to-end Gazebo simulation, we discarded any obstacle detections that were more than 5 meters away. 

\subsection{Recurrent Neural Networks for Learning Uncertainties}
\label{methods:rnn}
\label{methods:rnn-training-data}

\begin{figure*}[!t]
    \centering
    \includegraphics[width=0.95\textwidth]{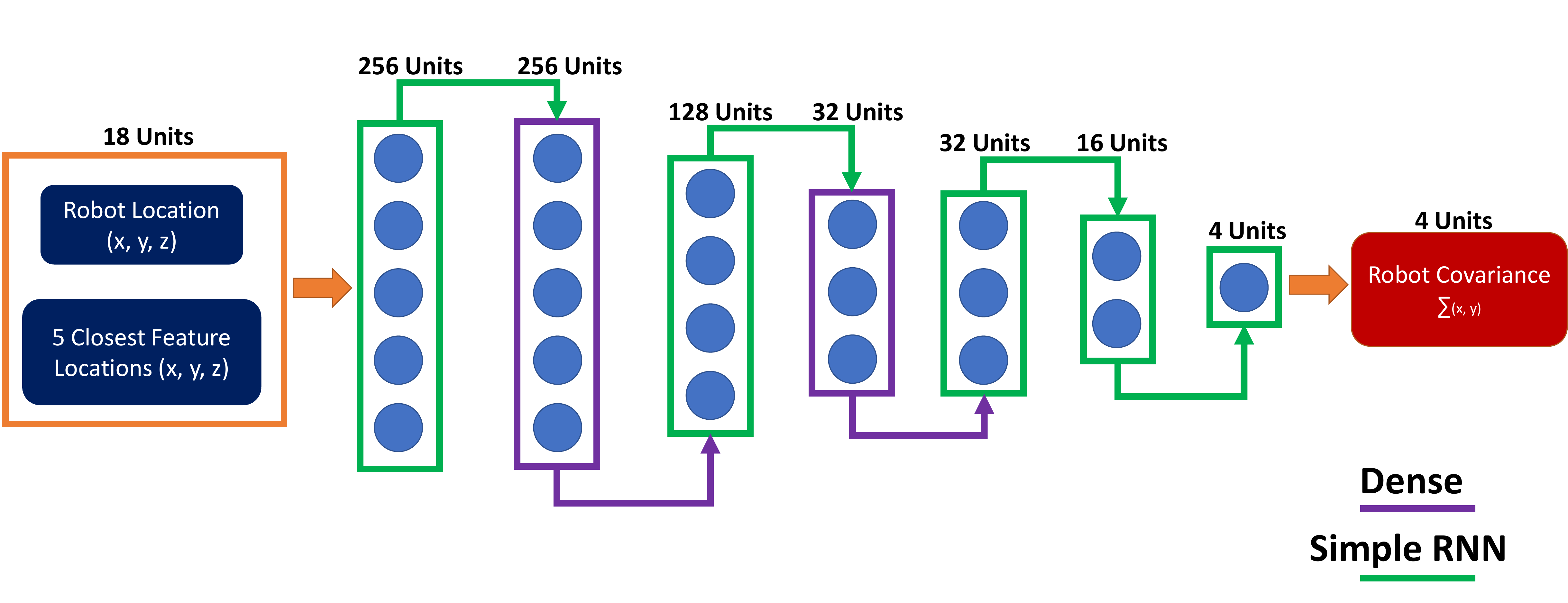}
    \caption{\textbf{Recurrent Neural Network Architecture.} Our RNN architecture predicts the covariances at robot poses [$x_{t+n}$, $y_{t+n}$] at timesteps $t+n$ for $n = 1,...,N$ (where $N$ is the length of the MPC's prediction horizon). During training, we used inputs collected from the output of XIVO to parameterize the network towards the four output units, as indicated by the first 18 input units and last four units in the figure above. Seven hidden layers were used with ReLU activation functions, with five recurrent layers (green) and two fully connected layers (purple), to learn the temporal structure for covariance propagation.}
    \label{rnn_arch}
\end{figure*}

The RNN, shown in Figure \ref{rnn_arch}, uses a combination of feedforward layers and simple RNN layers. The hidden layers all use ReLU activations. The network's 18 inputs are the robot's $x$, $y$, and $z$ positions. The next 15 inputs consist of the $x$, $y$, and $z$ positions of the five closest tracked features at any given state. The four output layer neurons correspond to the four values of the robot's \( 2 \times 2 \) $x$-$y$ covariance matrix, which is then used in Constraint V of the motion planning MPC. Unlike the hidden layers, the output layer uses a linear activation function because the outputs themselves are not restricted. Note that even though our MPC plans in only two dimensions, the inputs to the neural network are three-dimensional because the state estimation in our experiment is three-dimensional.

We used the Mean Squared Error (MSE) as the loss function:
\[ MSE = \frac{1}{N} \sum_{i=1}^{N}(\Sigma_i - \hat{\Sigma}_i)^{2} \]

\noindent Here $N$ is the total number of timesteps, \(\Sigma_i\) is the covariance matrix of the planer position computed by a SLAM system at timestep $i$ and \(\hat{\Sigma}_i\) is the covariance matrix predicted by the RNN. Conceptually, the covariance matrix is a \( 2 \times 2 \) matrix, but the implementation of the RNN treats it as a \( 4 \times 1 \) flattened matrix when it makes predictions and propagates error. Lastly, we note that covariance matrices are positive semidefinite by definition. However, the above training procedure does not constrain the output of the RNN to positive semidefinite; the outputs were indeed arbitrary 2 x 2 matrices. To account for this, we zeroed out off-diagonal elements and negated any negative diagonal elements.

Training data (robot position, position of tracked features, and covariance matrices) for the RNN was collected from running XIVO,\footnote{Code available: \url{https://github.com/ucla-vision/xivo}} a simplified and modernized implementation of the SLAM system described in \cite{jones_visual-inertial_2011}, on time-synchronized RGB and IMU data collected from an Intel RealSense D435i mounted onto ALPHRED's head. We collected four {\raise.17ex\hbox{$\scriptstyle\mathtt{\sim}$}}40-second training sequences in total (using 100 epochs for training on the four sequences). The right side of Figure \ref{cnn_box} displays tracked features and localization estimates from the collected data.

One key assumption that XIVO makes is that disturbances to the angular velocity and acceleration measurements (bias + noise) are a random walk (i.e. white, zero-mean, and Gaussian). This is not true for a walking robot, where each step produces a large periodic disturbance. Thus, XIVO's generic motion model is best suited to a flying robot. However, to adapt XIVO for our quadruped, we limited the acceleration and angular rate measurements to "realistic" values and then "de-tuned" the filter by setting large bounds on expected IMU measurement noise. This hampered accuracy, but ultimately enabled convergence.

%============================================================
% Experimental Results
%============================================================
\section{Experimental Results}

\begin{figure}[!t]
    \centering
    \includegraphics[width=0.95\columnwidth]{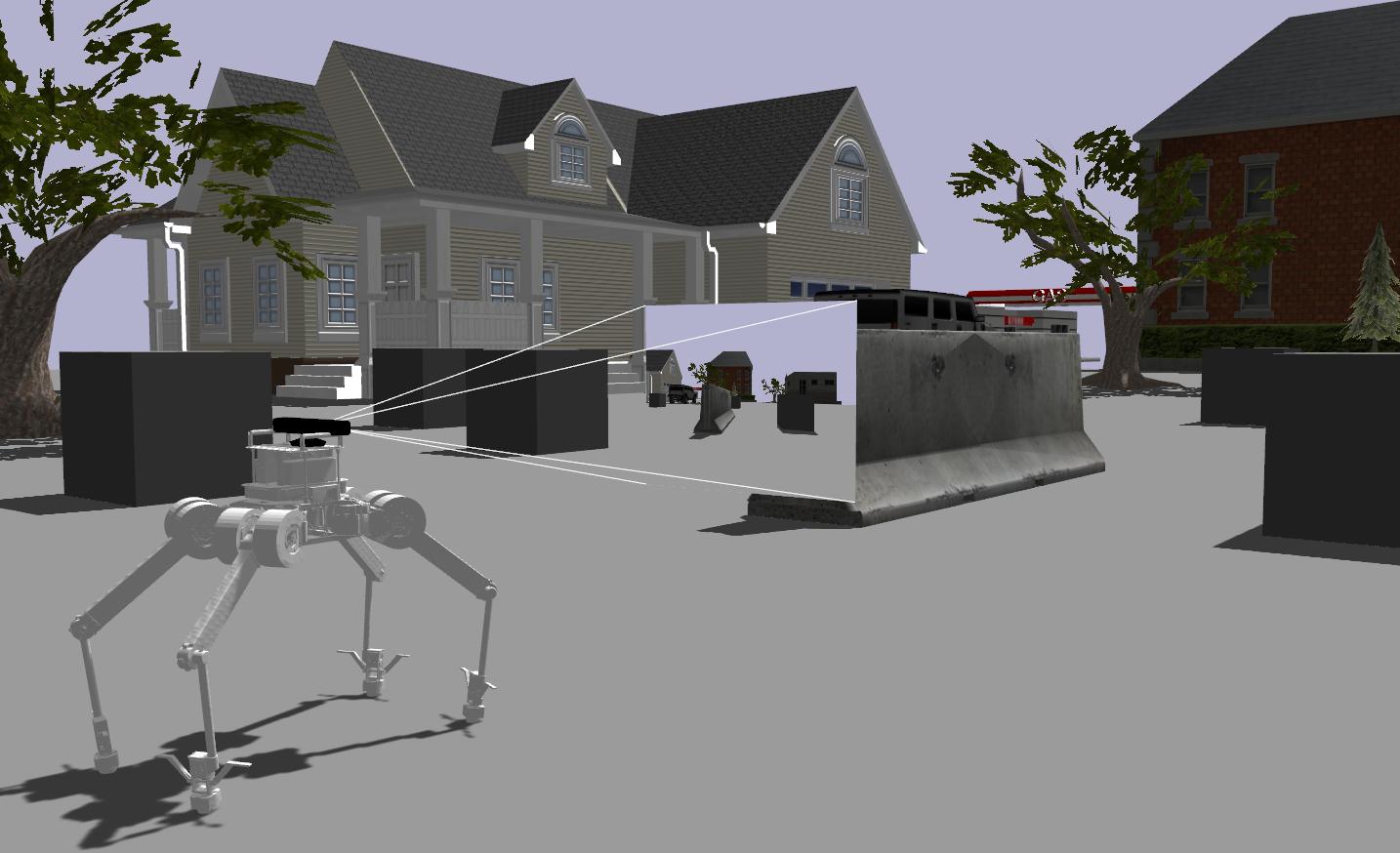}
    \caption{\textbf{Gazebo Simulation.} Our high-fidelity simulation accurately models the dynamics of the ALPHRED quadruped robot.}
    \label{gazebo_sim}
\end{figure}

In this section, we provide an overview of our robot model and its motion tracking controller, describe the training and testing results of the CNN and RNN neural networks, and then summarize our end-to-end results using our Gazebo simulation environment.

\subsection{Robot Model and Motion Tracking Controller}
\label{experimental results:A}
\begin{figure}[!t]
    \centering
    \includegraphics[width=0.95\columnwidth]{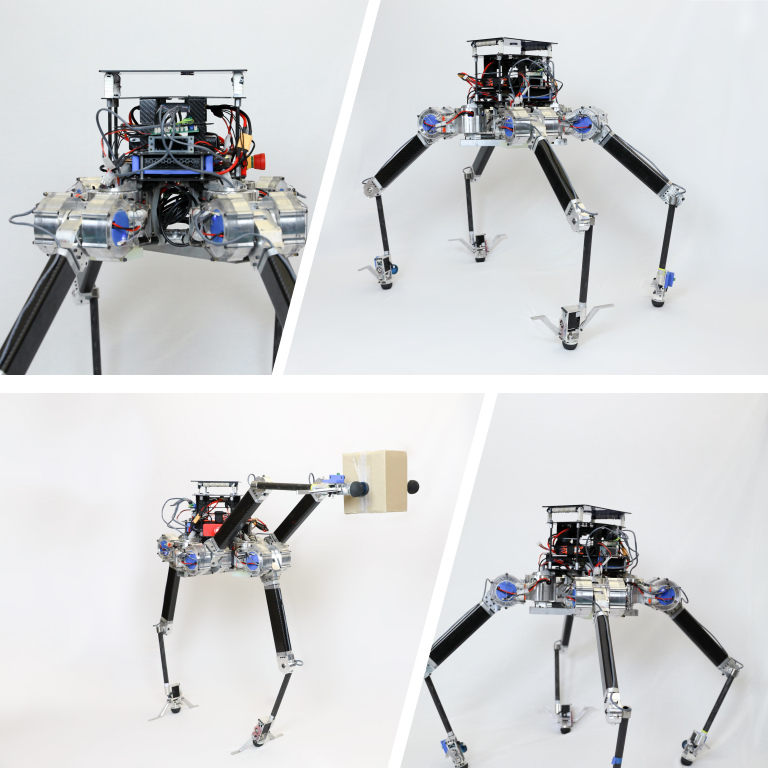}
    \caption{\textbf{ALPHRED Hardware.} The ALPHRED quadrupedal robot developed by Hooks \textit{et al.} \cite{hooks2020ALPHRED} of the RoMeLa robotics laboratory at the University of California, Los Angeles. This complex platform is an ideal model to apply our methods, as showing success on this platform also demonstrates the potential of applying our methods to a wide selection of robotic systems. Table~\ref{tab:robot_model} describes some physical properties of the system.}
    \label{fig_alphred}
\end{figure}

The robot used in this study is ALPHRED from Hooks \textit{et al.} \cite{hooks2020ALPHRED}, a full-sized quadrupedal robot that has unique kinematic configurations which enable several dynamic modes of operation as shown in \figref{fig_alphred} and Table \ref{tab:robot_model}. Our path planner is tested on a highly accurate simulation of ALPHRED using Gazebo software \cite{gazebo} (\figref{gazebo_sim}). The robot is modeled as several interconnected rigid-bodies in PyBullet so that the state includes not only joint angular velocities, but sensor and actuator noise due to motor temperature. The camera model used is a standard perspective projection with the same intrinsics as the Intel RealSense camera used to collect RNN training data, but without distortions. ALPHRED uses an Extended Kalman Filter (EKF) that fuses kinematic encoder data with on-board IMU measurements to provide full state estimation \cite{bloesch2013state}. A Raibert-style controller \cite{raibert1986legged} is used to track desired trajectories, where the input to the controller is desired planar velocities ($v_d$) and a desired yaw rate ($\dot{\psi}_d$) in the body frame. The controller operates by planning footsteps using powerful heuristics based on velocity feedback and corrects velocity and orientation errors by adjusting the length of the limbs in support. Further details of the ALPHRED platform and its low-level motion tracking controller can be seen in \cite{hooks2020ALPHRED}.

\subsection{Analysis of Learning Components}

\begin{figure}[!t]
    \centering
    \includegraphics[width=0.95\columnwidth]{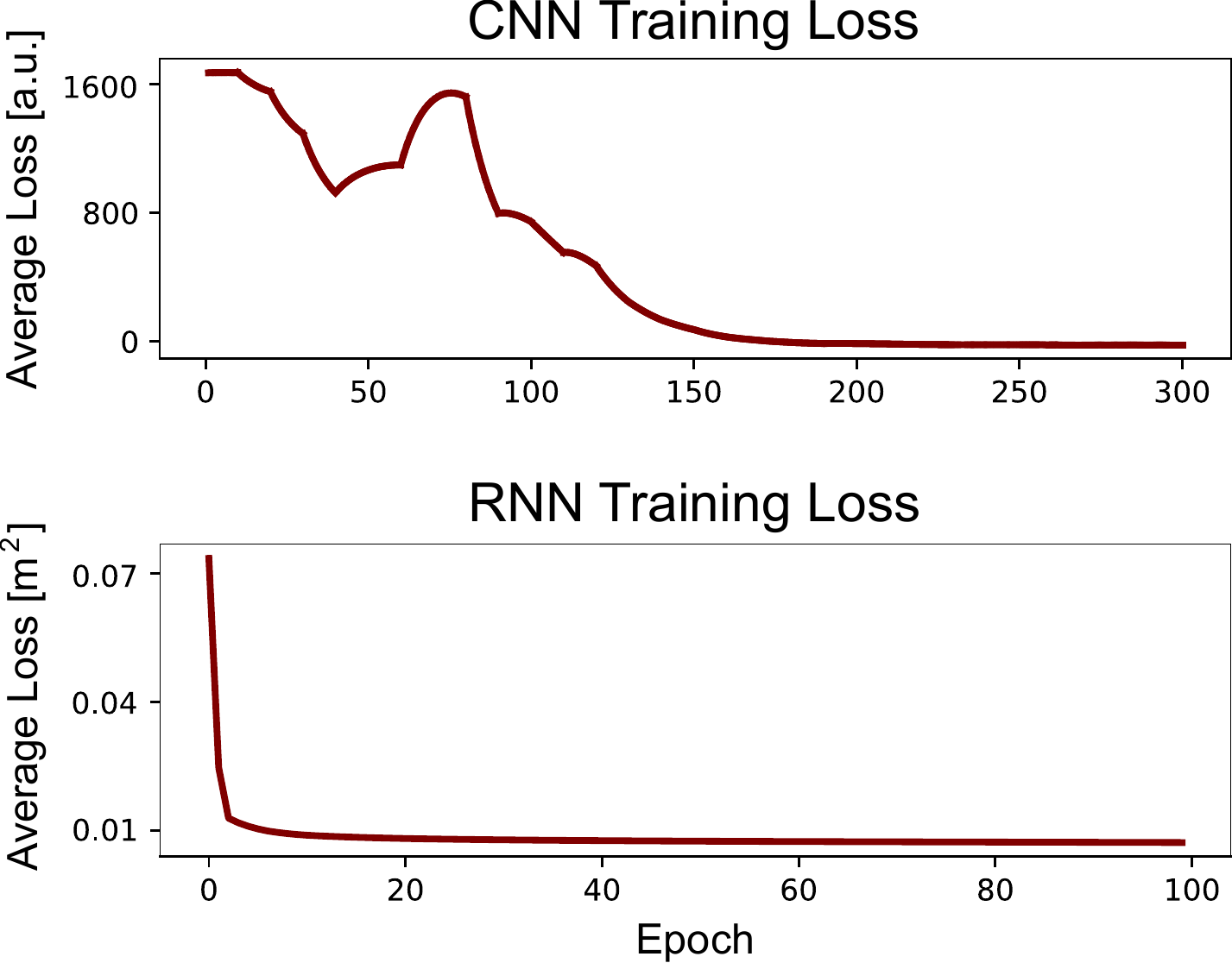}
    \caption{\textbf{Training Loss.} \textit{Top:} Our CNN model's training loss, used in our object detection pipeline. We trained for 5,200 epochs but only display 300 in the figure above. Note that we verified avoidance of overfitting via a validaton set but did not plot the curve here. \textit{Bottom:} Our RNN model's training loss, used to infer future localization uncertainty for the MPC. As with the CNN, we verified avoidance of overfitting using a validation set.}
    \label{fig:loss}
\end{figure}

Training loss for both the CNN and RNN are shown in Figure (\figref{fig:loss}). CNN and RNN networks were trained for 5,300 and 100 epochs, respectively, but only a limited range was plotted for visualization. To avoid overfitting, we used cross-validation and ensured that the validation loss was close to the training loss during the training process for both networks. Additionally, we observed that as ALPHRED tracked more features (i.e., the corners of an obstacle), the RNN's covariance estimates decreased. Conversely, as tracked features went out of view, estimates would increase. This is expected from the behavior of a visual-inertial odometry algorithm.

\subsection{Gazebo Simulation}

To test our proposed method, we used a custom Gazebo environment loaded with a high-fidelity model of our quadrupedal robot equipped with a Microsoft Kinect sensor. For localization, we used the motion tracking controller as described in Section~\ref{experimental results:A}. Our 3D environment consisted of a $1 m^3$ box obstacle with the objective to command ALPHRED to move from its initial position at [0,0] to the goal position at [8,0]. We compared our method against a baseline approach, in which only an MPC was used for trajectory planning (with the obstacle explicitly hardcoded), and a naive approach for safer traversal, in which the robot's radius was artificially inflated to twice the original size (from 0.7m to 1.4m).

\begin{figure}[!t]
    \centering
    \includegraphics[width=0.95\columnwidth]{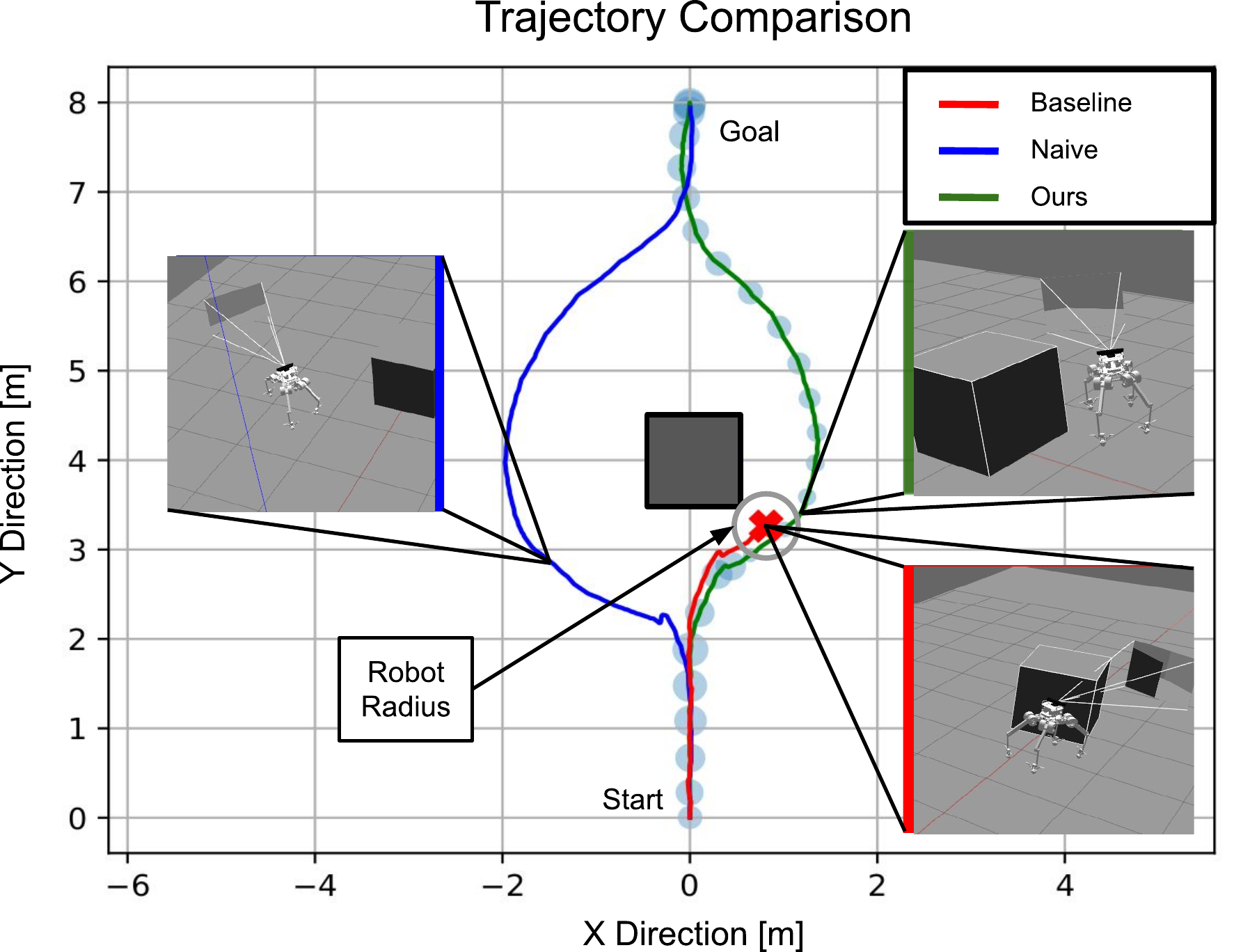}
    \caption{\textbf{Trajectory Comparison.} A comparison of the trajectories computed by three different approaches. The baseline method (red) is an MPC framework without our extensions to consider propagated future state uncertainty from an RNN, and we define the naive approach (blue) as artificially inflating a robot's boundary through all time. In comparison, our approach (green) can plan for a quick yet safe trajectory by predicting potential future collisions.}
    \label{fig:gazebo_trajectory}
\end{figure}

In the illustrative example shown in \figref{fig:gazebo_trajectory}, we observed that when using a classic MPC controller, which assumes that the robot's state estimation is perfect, the resulting trajectory is too close to the obstacle and ALPHRED crashes (red). On the other hand, when using a conservative MPC controller, in which the assumed value of ALPHRED's radius is twice its actual size, the resulting trajectory over-avoids collisions and ALPHRED moves slowly towards the goal point (blue). However, when using our full risk-aware MPC in this scenario, we observed that ALPHRED not only avoids collision, but executes a tighter trajectory than the conservative approach and requires less time to move to the goal. Note that the simulation was run on a laptop with an Intel Core i7 6700 HQ CPU and a NVIDIA GeForce GTX 970M GPU in real time with $dt_{plan} = 0.1s$ and $dt_{track} = 0.005s$.

%============================================================
% Discussion
%============================================================
\section{Discussion and Future Work}

% what is the takeaway of our work? think about how we can get people to cite this work. how can they use what we found in their future research? what do we contribute to the scientific community?

Collision-free path planning within unknown and unexplored environments requires the daunting integration of several components, such as sensor processing, control algorithms, and uncertainty resolution, into a fast and online end-to-end framework. To this end, we propose an architecture that unifies these modalities which attempts to address the fundamental problem of uncertainty in Active SLAM. By inferring the future positional uncertainty for an MPC using an RNN, we can substitute typical belief space planners with a more computationally efficient approach. Our work can also pave the way towards using RNNs to address problems with temporal structure which are difficult for classic robotic algorithms.

%In this paper, we introduce a fast end-to-end path planning architecture that attempts to address the fundamental problem of uncertainty in Active SLAM. Our approaches places several puzzle pieces 
Overall, our architecture addresses Active SLAM by combining MPC, SLAM, RNN, and CNN algorithms. We demonstrate that by inferring future positional uncertainties of the robot using our RNN prediction model, the robot can reach a goal state faster than when assuming a fixed uncertainty while still safely avoiding obstacles. This is significant because modeling uncertainties within a neural network framework, rather than belief space planning (i.e., POMDP), sufficiently shortens the computation time for hardware implementation. Future work will entail improvement of individual components in our architecture and modifying parts for complete hardware compatibility if necessary. For example, we would combine the classical feature detection and ALPHRED's state estimation (described in Section \ref{methods:obstacle-keypoint}) into a VIO algorithm.\footnote{We prefer a VIO algorithm over other SLAM algorithms because they can directly observe scale and because range sensors (e.g. LiDAR) are more expensive in both cost and computation and typically only produce sparse images. The motions of a walking robot should be sufficiently exciting, such that the VIO problem is observable.} This would require us to modify generic VIO equations, such as the ones present in XIVO, by explicitly modeling a walking gait, expanding the size of the gait space, and recomputing the Jacobians to incorporate robot dynamics into visual modeling instead of assuming a simple random walk. Then the performance-limiting detuning and signal clipping described in Section~\ref{methods:rnn-training-data} will become unnecessary. Finally, we also aim to replace the CNN + classical feature detection and unprojection with a modern semantic mapping algorithm, such as \cite{xiao-semantic-mapping}. 

Future directions also include: (1) formulating our RNN to infer semantics and feed object-dependent margins to the planner (e.g., the robot can get close to safe objects but not to dangerous ones); (2) exploring additional inputs to the RNN, such as the estimated covariances of features or other states; (3) incorporating a $z$ state/control in the MPC rather than assume planar motion for more complex path planning; (4) comparing our formulations to belief space planners (e.g., stochastic MPC) as well as other state-of-the-art path planners; and (5) constraining the RNN training process such that outputs are guaranteed positive semidefinite. We believe that implementing the above modifications could lead to closing the gap in achieving the futuristic vision for complete autonomous robotic systems.

\begin{table}[!t]
    \centering
    \caption{ALPHRED Configuration}
    \begin{tabular}{l|l}
    \bf Parameter & \bf Value \\ \hline
        Degrees of Freedom & 12 (3 per leg) \\
        Weight & 17.9 kg \\
        Max Velocity & 1.5 m/s \\
        IMU & VectorNav 200 \\ 
        Camera & RealSense D435i
    \end{tabular}
    \label{tab:robot_model}
\end{table}

\vfill\null

%============================================================
% References
%============================================================
\bibliographystyle{IEEEtran}
\bibliography{references}

% *****************************************************************
% 							   END 							    
% *****************************************************************
\end{document}